\documentclass{article}

\PassOptionsToPackage{numbers,square}{natbib}
\usepackage[final]{nips_2016}

\usepackage[utf8]{inputenc} 
\usepackage[T1]{fontenc}    
\usepackage{hyperref}       
\usepackage{url}            
\usepackage{booktabs}       
\usepackage{amsfonts}       
\usepackage{nicefrac}       
\usepackage{microtype}      

\usepackage[colorinlistoftodos]{todonotes}
\usepackage{amsmath}
\usepackage{xspace}
\usepackage{float}
\usepackage{subcaption}

\newcommand{\xvec}{\boldsymbol{x}}
\newcommand{\xmat}{X}

\newcommand{\eat}[1]{}

\title{Intra-day Activity Better Predicts Chronic Conditions}

\author{
Tom~Quisel\\
Evidation Health\\
Santa Barbara, CA\\
\texttt{tquisel@evidation.com} \\
\And 
David~C.~Kale \\
USC Information Sciences Institute\\
Los Angeles, CA\\
\texttt{kale@isi.edu} \\
\And 
Luca~Foschini\\
Evidation Health\\
Santa Barbara, CA \\
\texttt{lfoschini@evidation.com} \\
}

\begin{document}

\maketitle

\vspace*{-1em}

\begin{abstract}
In this work we investigate intra-day patterns of activity on a population of 7,261 users of mobile health wearable devices and apps. We show that: (1) using intra-day step and sleep data recorded from passive trackers significantly improves classification performance on self-reported chronic conditions related to mental health and nervous system disorders,
(2) Convolutional Neural Networks achieve top classification performance vs. baseline models when trained directly on multivariate time series of activity data, and (3) jointly predicting all condition classes via multi-task learning can be leveraged to extract features that generalize across data sets and achieve the highest classification performance.
\end{abstract}

\vspace{-1em}
\section{Introduction}

In recent years, the advent and rapid adoption of wearable technologies has made continuous ``life logging''~\citep{QS2016} a concrete possibility. As wearable technology matures, devices that track in-depth metrics such as minute-level step counts and minute-level sleep state are becoming more prevalent in medicine and health care, holding great promise for characterizing environmental and lifestyle factors that drive health outcomes outside the point of care.
Collectively referred to as ``mobile health'' (mHealth) these technologies are expected to play a significant role in enabling precision medicine, defined as ``treatment and prevention can be maximized by taking into account individual variability in genes, environment, and lifestyle''~\cite{precision2015report}. 
Despite recent initiatives~\cite{precision2015report} to create large cohorts of patients that share health data from a variety of sources including mHealth, the availability of population data with enough size and longitude to capture the weak signal that ties behavioral to longer-term health outcomes is still limited~\citep{fitbitROI2016}.

Especially unexplored is the impact that different timescales of data collection have on the inferences that can be derived from it. At one extreme, research has shown that data on millisecond timescales, such as that available from inertial sensors, allow for correctly identifying the type of human activity that is being performed at any point in time~\citep{lara2013survey}. At the other end of the spectrum, it is known that lower-frequency data, such as average daily-level step counts directly translate to risk factors for some chronic conditions~\citep{chiuve2006healthy}. 
Higher time resolution translates to more expensive sensors, shortened battery life, and higher storage needs. Therefore, a better understanding of the trade off of collecting data at a higher time resolution is warranted.




In this work we analyze activity recorded by passive trackers to investigate whether patterns that emerge at the minute level, but not at the daily level, can improve the characterization of someone's health state.
We present a Convolutional Neural Network (CNN) architecture trained on the multivariate time series of activity data at different time scales to infer an individual's self-reported chronic conditions. We show that our models significantly benefit from data at higher time scales and quantify the incremental contribution of activity data with at successively increasing time granularity up to minute-level.
Additionally, we apply multi-task training~\citep{ramsundar2015massively} across all chronic condition tasks to overcome noise and dimensionality challenges and show that it improves overall classification performance.

\vspace{-1em}
\section{Data}
\label{sec:data}

Users of a commercial reward-based wellness platform provided data for this study. Each user self reported demographic data, basic health metrics, indicated diagnosed conditions via an IRB-approved online health survey, and uploaded a recent history of step/weight/sleep data from passive activity trackers. We include all users who responded completely to the survey, reported at least 10 days of step data during the data collection window between 5/8/2016 and 10/1/2016, and reported at least one day of step and sleep minute-level data. Overall, 7,261 patients qualify of the 9,486 who responded to the health survey. The users in the data set are 75\% female with an average age of 38 years.

In the survey, each user reported a binary label for whether they had been diagnosed with each condition. We use a subset of these labels as the target labels for the classification tasks in this study. We focus on two clusters of conditions: a mental health/nervous system (MH/NS) cluster composed of 6 conditions: anxiety, depression, other mental illness, chronic pain, insomnia, and sleep apnea and a metabolic/circulatory (M/C) cluster composed of 3 conditions: hypertension, type 2 diabetes, and dyslipidemia. 

We stratify the input data for the classification task based on its time granularity. The layers are as follows: \texttt{demographic} (age, gender, ethnicity, education level, and parental status), \texttt{basic health} (weight, max weight in the past, height, BMI), \texttt{day-level activity} (per-day step counts, sleep durations, weight measurements, and binary utilization indicators for step, sleep and weight devices), \texttt{minute-level step} (31 per-day summary statistics computed from minute-level step data), and \texttt{minute-level sleep} (18 per-day summary statistics computed from minute-level sleep data). Example summary statistics are shown in Table~\ref{tab:features}. All values were passively recorded by the relevant tracker (i.e., pedometer, sleep trackers, scale) over the 147 day data collection window; none is self-reported.

Gaps in per-day values are imputed using per-user linear interpolation after censoring suspected partially-reported values. We find that partially-reported values are likely on the day immediately preceding and the day immediately following a tracking gap.

\begin{table}[ht]
\caption{Example per-day summary statistics computed from minute-level data}
\centering
\begin{tabular}{ll}
\toprule
Kind & Description\\
\midrule
Step & Length of longest streak with mean steps above 30 steps / minute\\
Step & Time of day for the first step taken\\
Step & Max step count over all 6 minute windows during the day\\
Sleep & Number of restless sleep periods during the night\\
Sleep & Time user goes to bed\\
\bottomrule
\end{tabular}
\label{tab:features}
\end{table}


\vspace{-1em}
\section{Methods}
We pose our predictive modeling problems as \textit{sequence classification} tasks. Given multivariate time series $\xmat = \left[ \xvec_1, \dots, \xvec_T \right]$ of tracked behavioral data for $T$ days for a user, we estimate the conditional probability $p(y\ |\ \xmat)$ of the target $y$ (e.g., a binary label indicating an anxiety diagnosis). In our full data set, $\xvec_t \in \mathbb{R}^{74}$ includes all per-day features as well as all static features (such as gender). The CNNs were implemented and trained using the open source Keras package ~\citep{chollet2015keras}.

\subsection{CNN architecture}
\begin{figure*}
\includegraphics[width=\textwidth]{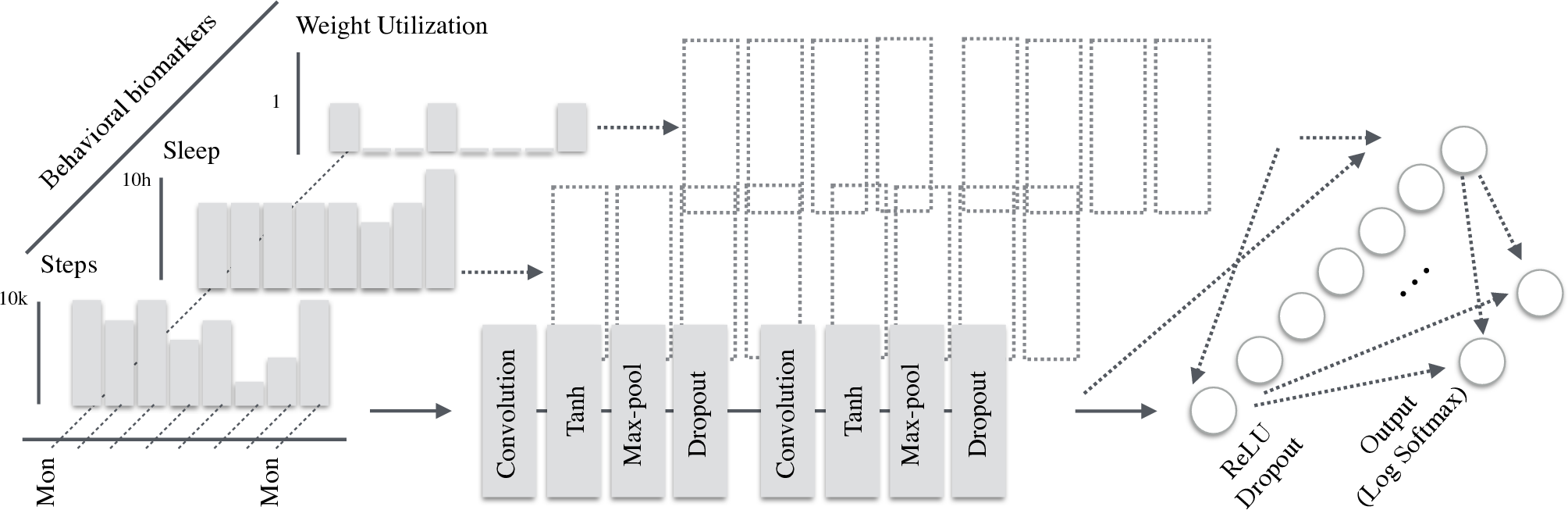}  
\caption{The temporal CNN architecture.}
\label{fig:CNN_arch}
\end{figure*}

Our temporal convolution neural net architecture is shown in Figure~\ref{fig:CNN_arch}.
Each individual sequence $X_u^{(k)} = [ x^{(k)}_1, \dots, x^{(k)}_T]_u$ (e.g., step counts) is fed separately to a two-stage univariate feature extractor, where each stage consists of a 1D temporal convolution followed by a non-linear activation function and a pooling operation. In contrast to previous work on time series CNN~\citep{zheng2014time}, we use a hyperbolic tangent non-linearity ($\tanh$) and max-pooling (vs. a sigmoid and average pooling) and dropout of probability 0.5 before each pooling layer.
The output of the feature extraction layers is flattened and fed to a standard fully connected multilayer perceptron (MLP) with one hidden layer~\citep{lecun2012efficient}. The hidden layer uses a rectified linear (ReLU) activation function and dropout of probability 0.5 before the final sigmoid output. 
The CNN is trained using gradient descent with backpropagation to minimize the negative log likelihood of the true label $y$: $\mathrm{loss}(y,\hat{y}) = -\left(y \log\hat{y} + (1-y) \log (1-\hat{y})\right)$ where $\hat{y} = p(y\ |\ \xmat)$.

\subsection{Multi-task training}

Multi-task training can improve performance on individual tasks, especially in the absence of large labeled data sets and when the tasks are related~\citep{caruana1996using,LiptonKEW15}.
To train a single neural net to solve $C$ different predictive tasks simultaneously, we add a separate output (with its own output weights $\boldsymbol{w}_c$) for each task $c$. The training loss for a single example with label vector $\boldsymbol{y} = [y_1, \dots, y_C]$ is the average over the individual task losses $\mathrm{loss}(\boldsymbol{y}, \boldsymbol{\hat{y}}) = (1/C) \sum \mathrm{loss}(y_c, \hat{y}_c)$. 

\vspace{-1em}
\section{Experimental results} 

We perform a series of binary classification experiments using the data sets and labels described in~\autoref{sec:data}. In all cases we predict the individual binary condition diagnosis labels for each user, there is no aggregation of labels to form tasks at the condition-cluster level. The CNN is trained in both a Single Task (ST) setting and a Multi-Task (MT) setting. All baseline models are trained only in the ST setting. The ST setting trains a separate model for each task (condition), while the MT setting trains a single model for all 9 tasks (6 MH/NS and 3 M/C) simultaneously.

We compare the performance of the CNNs and RNNs to a number of baseline models. These baselines include a linear logistic regression (LR) model with $L_2$ penalty trained on the raw time series input used for the NN models, an LR model trained on a set of hand-engineered features aggregating each time series into a single value, and a Random Forest (RF) using hand-engineered features.

We measure classifier performance using area under the ROC curve (AUC), averaged across four cross-validation folds. Each fold consists of a training (50\%), validation (25\%), and test (25\%) dataset. The classifiers are trained on the training set, hyperparameters are tuned on the validation set, and all AUCs were computed on the held-out test set. Significance tests are performed using paired t-tests across all tasks and folds.

Our final CNN architecture includes two convolutional layers of 8 and 4 filters with kernels of width 7 and 5, respectively. Both filters use step size of length 2, and are followed by max pooling with width 2 and step size 2. The fully-connected hidden layer has 300 nodes.

\subsection{Classifier comparison}

Classifier performance is summarized on the most detailed dataset (containing all data layers) for the two disease clusters in~\autoref{fig:classifier_auc}. Averaged across all 9 tasks from both clusters the MT CNN had the highest AUC at .719, significantly better (p=.01) than the ST CNN at .704, the RF at .700, the LR (Features) at .698, and the LR (Raw) at .642. The MT CNN achieves the highest overall AUC, .821, on the Type 2 Diabetes task. These results support the effectiveness of the MT approach for improving performance on related tasks over a medium-sized data set. Also notable is the superior performance of the CNNs over the LR (Raw) algorithm, demonstrating that CNNs can learn effective feature representations and avoid the need for hand engineering of features.

\subsection{Data layer comparison}

\begin{figure}
\centering
\begin{subfigure}{.5\textwidth}
  \centering
\includegraphics[width=.9\linewidth]{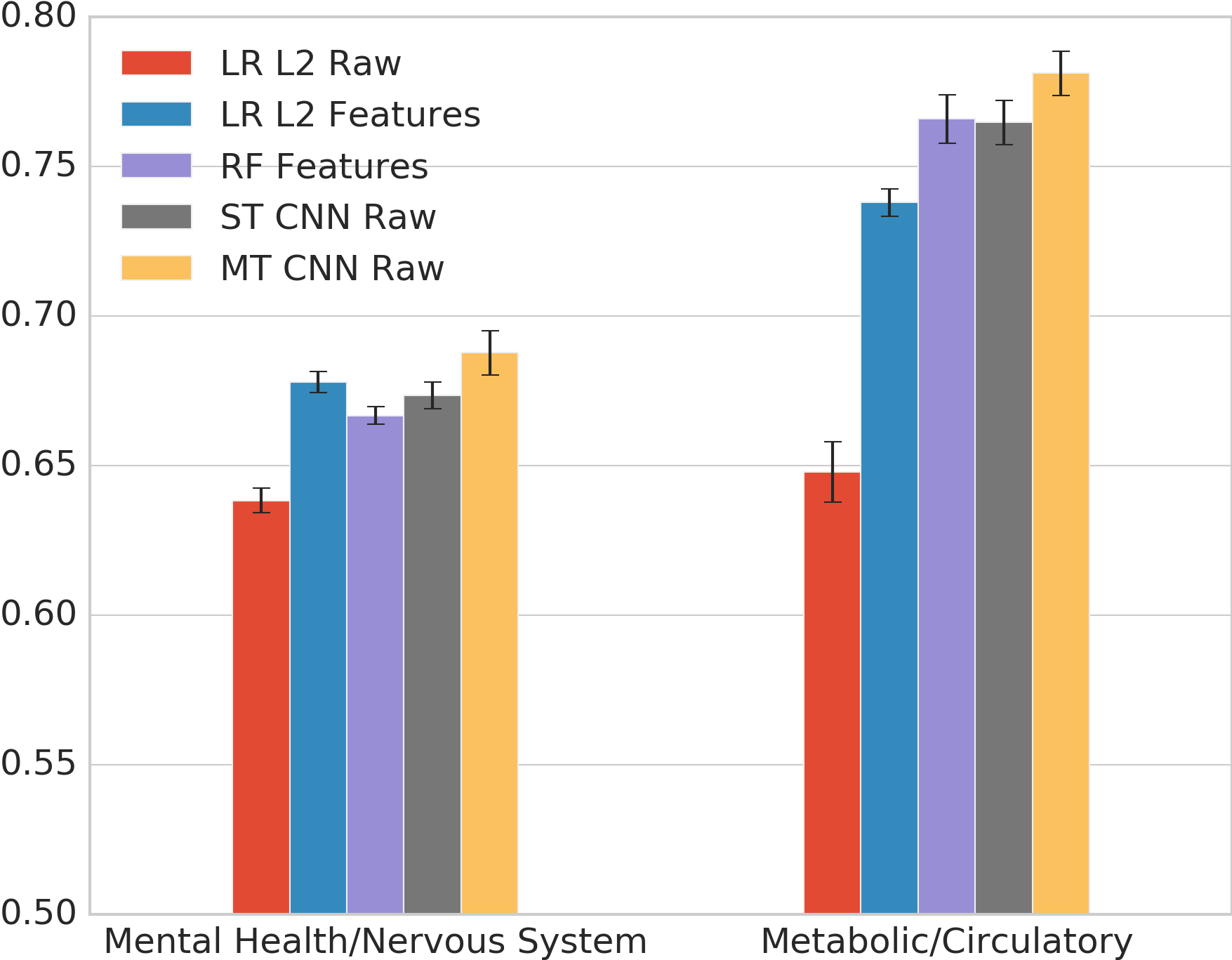}
\caption{AUC by classifier for the full dataset}
\label{fig:classifier_auc}
\end{subfigure}%
\begin{subfigure}{.5\textwidth}
  \centering
  \includegraphics[width=.9\linewidth]{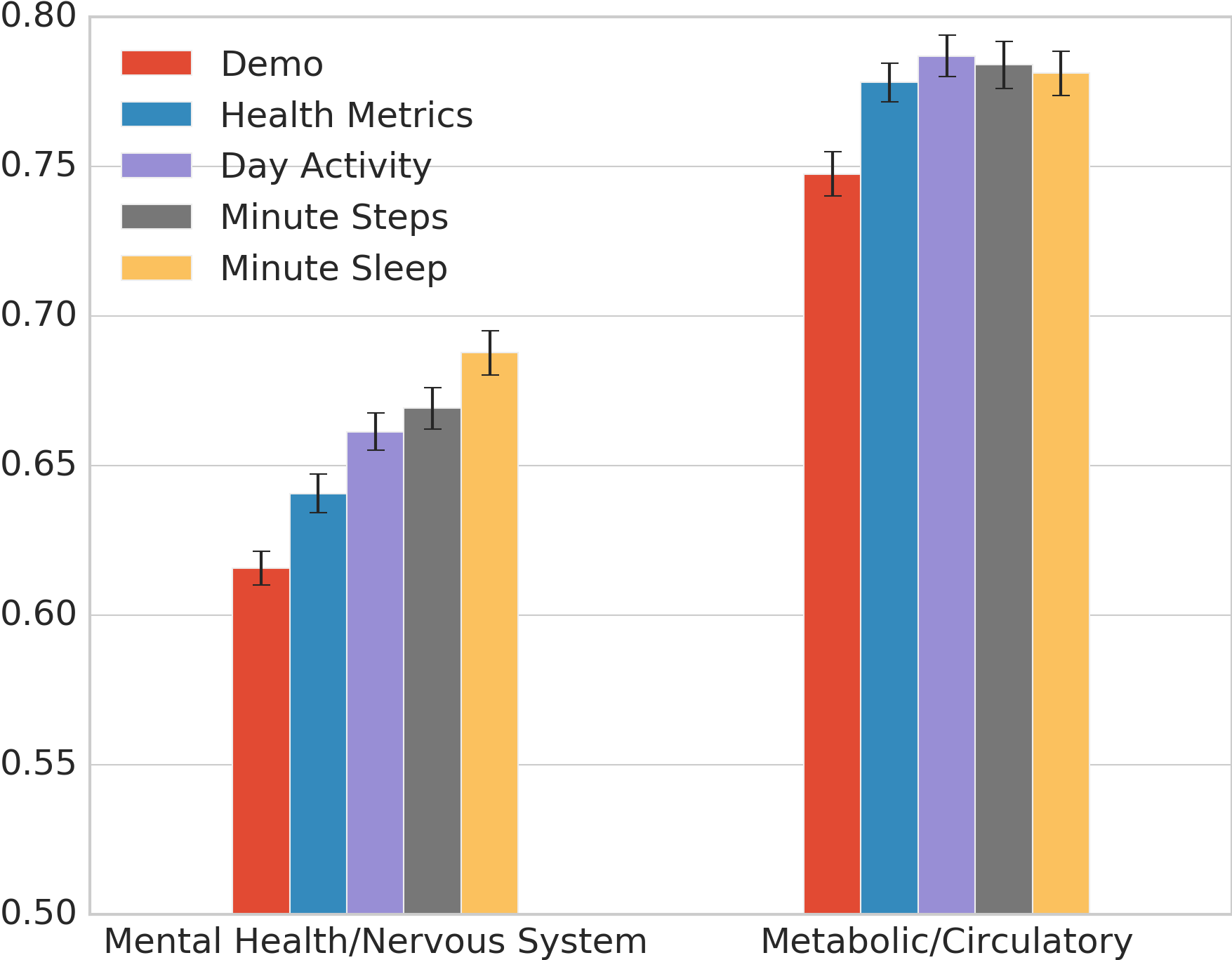}
  \caption{AUC by data layer for the MT CNN}
  \label{fig:data_layer_auc}
\end{subfigure}
\caption{Classification performance by condition cluster. Error bars are standard error of the mean.}
\label{fig:test}
\end{figure}

To measure the contribution of each additional layer of data to classification performance, we evaluate the the MT CNN across all 9 conditions using data at increasingly higher time granularity. We start with \texttt{demographic} data, then layer on \texttt{basic health}, \texttt{day-level activity}, \texttt{minute-level step}, and \texttt{minute-level sleep}. The results are summarized in~\autoref{fig:data_layer_auc}. Averaged across all tasks from both disease clusters, the respective AUCs are .660, .686, .703, .707, and .719. The addition of the activity-tracker layers provides the largest incremental AUC improvement for the MH/NS condition cluster. For this cluster, adding \texttt{day-level activity} improves the AUC by .021 (p<.001), \texttt{minute-level step} improves AUC by .008 (p=.022), and \texttt{minute-level sleep} data improves AUC by .019 (p=.003). Unlike  mental health/nervous system-related conditions, metabolic/circulatory conditions are more accurately predicted by \texttt{demographic} and \texttt{basic health} data only.

\vspace{-1em}
\section{Discussion} 
This work represents one of the first large scale studies using intra-day activity data collected from commercial mHealth devices in uncontrolled settings to identify self-reported medical chronic conditions. 
We show that activity data at a higher time granularity adds value for identifying some medical conditions related to mental health and disorder of the nervous system. For such conditions, we find that: (1) day-level step/sleep/weight data, (2) minute-level step, and (3) minute-level sleep data each contribute a statistically significant improvement in classification AUC for the task of identifying self-reported conditions for each user. On the contrary, we note that for conditions related to disorders of the metabolic and circulatory system, the performances of our models does not benefit from data at increased time granularity. 
We believe that these findings can inform the design of sensing frameworks and data analysis strategies tailored to different clinical contexts.
Our results show that CNN trained on raw temporal time-series attain good  performance, which is enhanced when tasks are predicted jointly via multi-task learning~\citep{ramsundar2015massively}. We surmise that the MT setting allows the CNN to learn feature representations over more common conditions and then leverage those representations when classifying rarer conditions. We believe that our findings constitute a step towards enabling more personalized solutions in medicine and health care~\citep{hood2011predictive}.

\bibliographystyle{abbrvnat}
\bibliography{main}


\end{document}